%% file: main.tex
\documentclass{article}

\usepackage[numbers]{natbib}
\usepackage[preprint]{neurips_2025_custom}

\input{math_commands.tex}

\usepackage[dvipsnames]{xcolor}         
\definecolor{linkColor}{rgb}{0.18,0.39,0.62}
\usepackage[utf8]{inputenc} 
\usepackage[T1]{fontenc}    
\usepackage[colorlinks=true,linkcolor=linkColor,citecolor=linkColor,filecolor=linkColor,urlcolor=linkColor]{hyperref}       
\usepackage{cleveref}
\usepackage{multirow}
\usepackage{xspace}
\usepackage{booktabs}
\usepackage{graphicx}
\usepackage{array}
\usepackage{wrapfig}
\usepackage{bm}
\usepackage{algorithm}
\usepackage{algpseudocode}
\usepackage{enumitem}
\usepackage{tcolorbox}
\usepackage{makecell}
\usepackage{diagbox}

\usepackage{amsmath}
\usepackage{amssymb}
\usepackage{mathtools}
\usepackage{amsthm}

\input{settings.tex}

\newcommand{\github}{\raisebox{-1.5pt}{\includegraphics[height=1.05em]{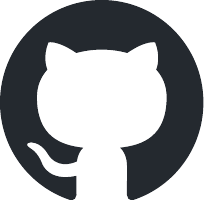}}\xspace}
\newcommand{\house}{{\raisebox{-1.5pt}{\includegraphics[height=1.05em]{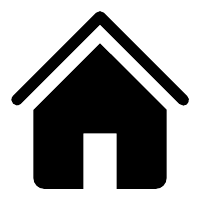}}}\xspace}

\definecolor{DarkBlue}{RGB}{0, 51, 153}
\newcommand{\ours}{GAD}

\title{Black-Box Knowledge Distillation of Proprietary Large Language Models}
\title{Black-Box Knowledge Distillation of Large Language Models}
\title{Generative Adversarial Distillation: \\ Black-Box On-Policy Distillation of Language Models}
\title{Black-Box On-Policy Distillation of \\ Large Language Models}

\author{%
Tianzhu Ye\thanks{~Equal contribution. Contact person: \href{mailto:fuwei@microsoft.com}{fuwei@microsoft.com}.}~~~~~~~~Li Dong\footnotemark[1] \\
~\bf Zewen Chi~~~~~~Xun Wu~~~~~~Shaohan Huang~~~~~~Furu Wei \\
~Microsoft Research \\
~{\href{https://aka.ms/GeneralAI}{https://aka.ms/GeneralAI}}
}

\begin{document}

\maketitle

\begin{abstract}
Black-box distillation creates student large language models (LLMs) by learning from a proprietary teacher model's text outputs alone, without access to its internal logits or parameters.
In this work, we introduce \textbf{G}enerative \textbf{A}dversarial \textbf{D}istillation (\ours{}), which enables on-policy and black-box distillation.
\ours{} frames the student LLM as a generator and trains a discriminator to distinguish its responses from the teacher LLM's, creating a minimax game.
The discriminator acts as an on-policy reward model that co-evolves with the student, providing stable, adaptive feedback.
Experimental results show that \ours{} consistently surpasses the commonly used sequence-level knowledge distillation.
In particular, Qwen2.5-14B-Instruct (student) trained with \ours{} becomes comparable to its teacher, GPT-5-Chat, on the LMSYS-Chat automatic evaluation.
The results establish \ours{} as a promising and effective paradigm for black-box LLM distillation.

\end{abstract}

\vfill

\begin{table}[H]
\centering
\begin{tabular}{@{}r@{\hspace{2pt}}l@{}}
\house & \textbf{Project Page}: \href{https://aka.ms/GAD-project}{\texttt{aka.ms/GAD-project}}\\
\github & \textbf{Code}: \href{https://aka.ms/GAD-github}{\texttt{aka.ms/GAD-github}}
\end{tabular}
\end{table}

\vfill

\begin{figure}[H]
\centering
\includegraphics[width=0.95\textwidth]{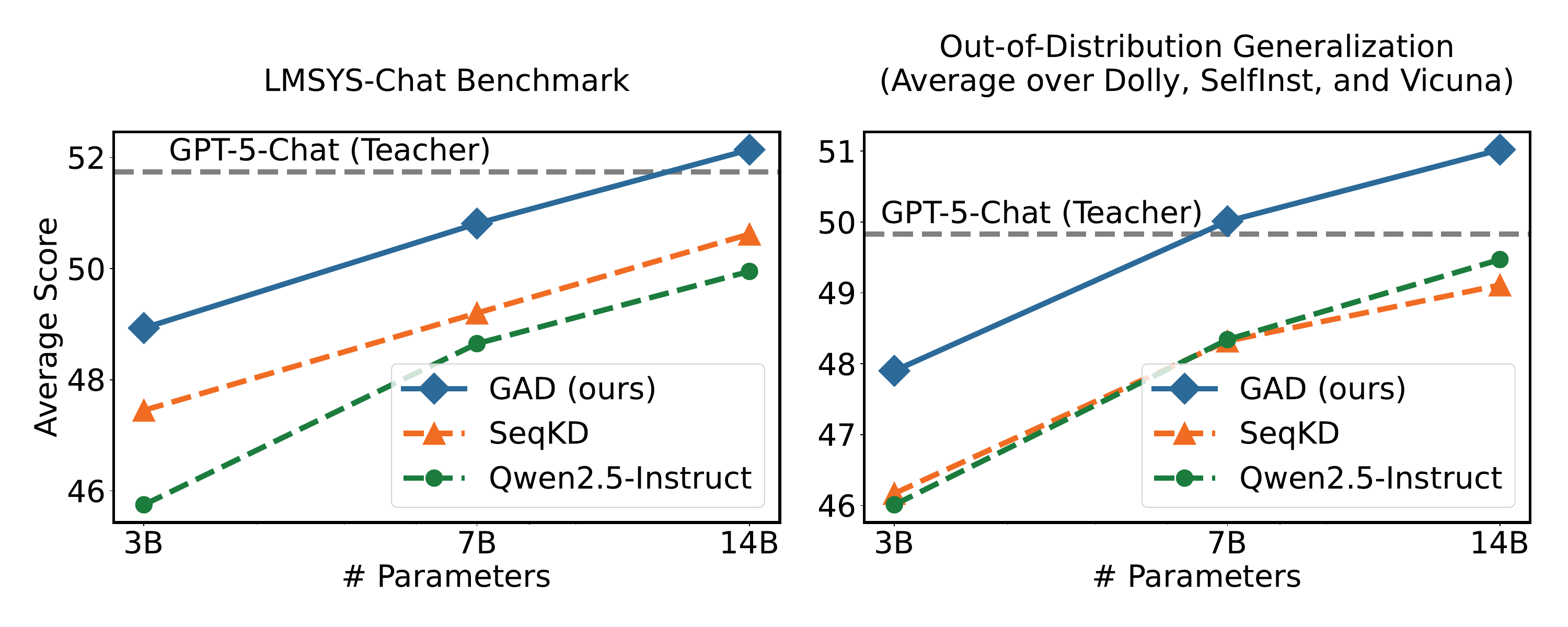}
\caption{Comparison between \ours{} and sequence-level knowledge distillation (SeqKD;~\citealp{skd}) trained on LMSYS-Chat~\citep{lmsys} dataset, evaluated by averaged GPT-4o scores. \textbf{Left}: Results on the LMSYS-Chat test set. \textbf{Right}: Average performance across Dolly~\citep{dolly}, SelfInst~\citep{self_inst}, and Vicuna~\citep{vicuna} datasets.}
\label{fig:scaling_student}
\end{figure}

\vfill

\newpage
\section{Introduction}

Knowledge distillation~\citep{kd} in large language models (LLMs;~\citealp{gpt4,gpt5,deepseekv3,qwen3}) is primarily used to create smaller, more efficient student models that retain much of the performance of a larger, resource-intensive teacher model.
The setting in which the student has access to the teacher's internal probability distribution or hidden states is called \textit{white-box distillation}.
Standard white-box approaches align the teacher and student by matching their output distributions, typically via Kullback-Leibler divergence (KLD)~\citep{lightpaff,minillm}, or their inner states~\citep{tinybert,bert-pkd,minilm}.
However, white-box access is often impractical when the teacher is a proprietary API model (e.g., GPT-5). In this scenario, only teacher-generated texts are accessible, defining the more challenging black-box distillation setting.
The absence of fine-grained probability supervision makes conventional likelihood-based objectives unavailable. Typical black-box distillation methods simply perform supervised fine-tuning on teacher responses~\citep{alpaca,vicuna}. Furthermore, when the student and teacher employ incompatible tokenizers, applying likelihood-based objectives also becomes challenging.
This highlights the need for a framework that can effectively extract deeper and richer knowledge from teacher-generated text responses.

Recent studies~\citep{minillm,googlepolicy,thinkingmachine-onpolicy,qwen3} in white-box distillation highlight the importance of \textit{on-policy} learning, where the student learns from its own generated responses rather than solely imitating the teacher's outputs.
These studies show that performing reverse KLD on student-generated text promotes mode-seeking behavior and reduces exposure bias compared to teacher-forced training.
However, extending this idea to the black-box setting introduces a major challenge: when the student produces its own responses, there are no probability-level supervision signals available from the teacher to evaluate or correct them.
Without explicit feedback, the student cannot directly gauge the quality of its generations relative to the teacher, making effective on-policy distillation infeasible under the standard likelihood-based framework.

To address this limitation, we propose \textbf{\ours{}}, a \textbf{G}enerative \textbf{A}dversarial \textbf{D}istillation framework that enables on-policy learning in the black-box regime.
Our key idea is to view the student as a \textit{generator} that produces responses conditioned on prompts, and to train a \textit{discriminator} to distinguish between teacher and student outputs.
The generator is then optimized to produce responses that the discriminator cannot distinguish from those of the teacher, forming a minimax game similar to generative adversarial networks (GANs;~\citealp{gan,seqgan}).
This adversarial process allows the student to receive implicit feedback on the quality of its own generations, even without access to the teacher's probability space.
Besides, from the perspective of reinforcement learning (RL;~\citealp{rlintro,ppo,trpo}), our discriminator can be interpreted as an \textit{on-policy reward model} that evolves jointly with the student policy.
Unlike conventional reward models in RLHF~\citep{instruct-gpt} which are fixed after pretraining and prone to reward hacking~\citep{reward_hacking}, our discriminator continually adapts to the student's behavior during training. The on-policy reward modeling provides stable and dynamic supervision throughout the distillation process.

We validate our approach using GPT-5-Chat~\citep{gpt5} as a teacher and a range of open-source models from the Qwen2.5~\citep{qwen2.5} and Llama3~\citep{llama3} families as a student.
Experiments are conducted on the a subset of LMSYS-Chat-1M dataset~\citep{lmsys} and evaluated across multiple domains.
Under identical training budgets, \ours{} consistently outperforms both the instruction models before distillation and the SeqKD~\citep{skd,vicuna,alpaca,ITGPT4,lima} baseline across all datasets and model sizes.
Notably, on GPT-4o score, Qwen2.5-3B-Instruct distilled with \ours{} matches the performance of Qwen2.5-7B-Instruct distilled with SeqKD, while Qwen2.5-14B-Instruct trained with \ours{} approaches the capability of the GPT-5 teacher itself.
Our method also delivers particularly strong improvements in out-of-distribution generalization, where SeqKD yields marginal or negative gains.
Human evaluations further confirm performance. \ours{} can effectively extract high-quality knowledge from black-box LLMs without access to output logits.

\newpage
\section{Method}
\label{sec:method}

We study conditional text generation of large language models, where a model generates a response $y$ conditioned on a given prompt $x$ sampled from dataset $\mathcal{T}$. 
To transfer the capabilities of large models to smaller ones, knowledge distillation (KD) trains a student distribution $q_\theta(y | x)$ parameterized by $\theta$ to approximate the behavior of a teacher distribution $p(y | x)$. 
In the white-box distillation setting, the student has access to the teacher's predictive distribution $p(y | x)$. Approaches such as forward KLD~\citep{skd,lightpaff,vicuna,alpaca} or reverse KLD~\citep{minillm} are designed for this setting. 
However, these techniques can fail if the teacher is a proprietary model that only returns generated text. We refer to this scenario as \textit{black-box distillation}, where only textual responses from the teacher are observable. The goal is to learn a student model that imitates the teacher's generative behavior without access to its internal probability space.

\subsection{\ours{}: Generative Adversarial Distillation}
\label{sec:gad}

\begin{figure}[t]
\centering
\includegraphics[width=\linewidth]{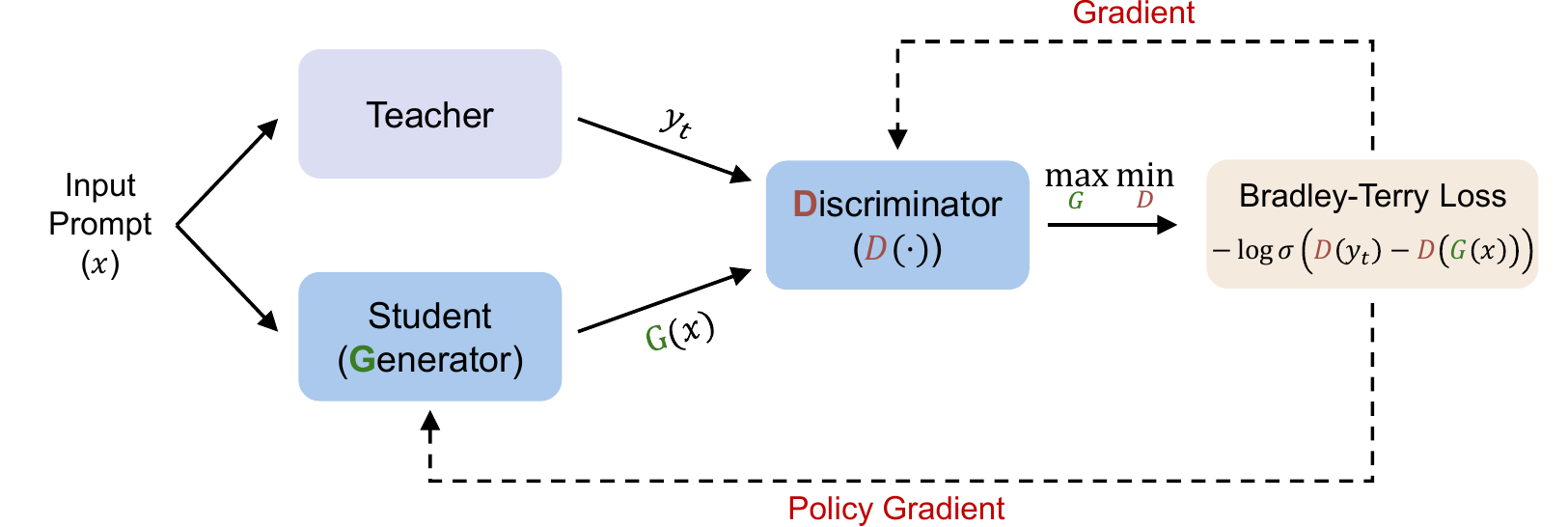}
\vspace{-0.1cm}
\caption{Training procedure of \ours{}. The student (generator) learns to generate responses that maximize the score assigned by the discriminator. The discriminator is trained with Bradley-Terry loss to assign a lower score to the student than the teacher, learning to distinguish between them. Together, they form a two-player minimax game in an adversarial learning framework.}
\vspace{-0.1cm}
\label{fig:method}
\end{figure}

We perform black-box distillation with generative adversarial training~\citep{gan,seqgan} as shown in \Cref{fig:method}.
The training dataset $\mathcal{T}=\{(x, y_t)\}$ is constructed by iterating over the prompts $x$ in the original dataset and sampling a teacher response $y_t$ for each.
Our framework consists of a generator $G$ which is the student model, and a discriminator $D$ that assesses the quality of the student and teacher responses.
The generator generates the response $G(x)$ to the prompt $x$.
The discriminator predicts a sequence-level scalar score $D([x,y])$ given prompt $x$ and response $y$\footnote{~The input prompt $x$ and generated response $y$ are concatenated (i.e., $[x,y]$) and fed into the discriminator (i.e., $D([x,y])$).  For brevity, we use $D(y)$ below to represent $D([x,y])$.}. The discriminator is initialized using generator model parameters with an extra prediction head. The head projects the final hidden state to a scalar score, and the score of the last token in the sequence is taken as the sequence-level score.
The training objective is formulated as a two-player minimax game with the following value function $\mathcal{V}(G, D)$:
\begin{equation}
\label{eq:minmax}
\max_G \min_D \ \ \mathcal{V}(G, D) = \mathbb{E}_{(x,y_t) \sim \mathcal{T}} \left[-\log \sigma\left( D(y_t) - D(G(x)) \right)\right],
\end{equation}
where $\sigma(\cdot)$ denotes the sigmoid function. We use Bradley-Terry model~\citep{bradley} to capture pairwise preferences between teacher and student response.
The proposed generative adversarial training framework allows the student to learn on-policy from its own generated responses via discriminator feedback, eliminating the need to access the teacher's internal representations.

\subsection{Training}
\label{sec:rl_optim}

We discuss the training algorithm of generator and discriminator respectively. From \Eqref{eq:minmax}, the generator $G$ is trained with the following objective:
\begin{equation}
\label{eq:generator}
\text{(\textit{Generator})} \quad ~~ \max_G \ \mathbb{E}_{(x,y_t) \sim \mathcal{T}}\left[D(G(x))\right],
\end{equation}
Since the sampling operation in $G(x)$ is non-differentiable with respect to the student model parameters, we treat $D(G(x))$ as a reward and optimize it using policy gradient~\cite{policy_gradient} with established reinforcement learning algorithms. We employ GRPO~\citep{grpo} to train the student in our experiments, with detailed formulations provided in Appendix~\ref{app:grpo}.
For the discriminator $D$, we minimize its training loss derived from \Eqref{eq:minmax}:
\begin{equation}
\label{eq:discriminator}
\text{(\textit{Discriminator})} \quad ~~ \min_D \ \mathbb{E}_{(x,y_t) \sim \mathcal{T}} \left[-\log \sigma\left( D(y_t) - D(G(x)) \right)\right].
\end{equation}
The discriminator uses Bradley-Terry loss to capture pairwise preferences, encouraging higher scores for teacher responses over student-generated ones.

\paragraph{Warmup Before \ours{} Training}
We find that jointly warming up the generator and discriminator before the \ours{} training stage is crucial for final performance.
We fine-tune the student on the teacher's response, and we minimize the cross-entropy loss as warmup for the generator.
In the meanwhile, the discriminator is trained using the same data with the Bradley-Terry loss in \Eqref{eq:discriminator}.
We conduct warmup for both models for one epoch before starting \ours{} training. This step promotes effective adversarial optimization and ensures the balance between the generator and discriminator. Ablation studies on the warmup strategy are presented in \Cref{sec:ablation}.

\subsection{Implement \ours{} with Reinforcement Learning Frameworks}
\label{sec:implementation}

In our experiments, we implement \ours{} using existing reinforcement learning frameworks, such as verl~\cite{verl}.
GRPO~\cite{grpo} is used as the policy gradient algorithm, which is detailed in Appendix~\ref{app:grpo}.

As presented in \Cref{tbl:rl_vs_gad}, we implement the generator as a \textit{policy model} and the discriminator as a \textit{reward model}.
The generator produces responses, receives rewards from the discriminator, and is optimized to maximize the expected reward.
The reward is defined in \Eqref{eq:generator}, i.e., $D(G(x))$.

Unlike vanilla reinforcement learning, \ours{} also needs to jointly update the discriminator (i.e., reward model).
The discriminator is trained with Bradley-Terry loss on preference pairs to score the teacher response higher than the student's output, similar to the reward model in RLHF~\citep{instruct-gpt}. While conventional RLHF trains a fixed reward model prior to policy optimization which is prone to reward hacking, our approach updates the reward model (discriminator) online to adapt it to the current policy continually.

\begin{table}[ht]
\centering
\begin{tabular}{l m{0.32\textwidth} m{0.32\textwidth}}
\toprule
 & \textbf{Reinforcement Learning} & \textbf{GAD} \\
\midrule
\multirow{3}{*}{\tabincell{l}{\bf Term \\ \bf Correspondence}} & {Policy Model} & {Generator (i.e.,Student LLM)}  \\
 & {Reward Model} & {Discriminator}  \\
 & {Reward} & {$D(G(x))$ (as in \Eqref{eq:generator})}  \\
\midrule
\textbf{Difference} & The reward model is typically trained once on a static dataset and then \textbf{frozen}. The policy is then optimized against this fixed reward function. & The discriminator \textbf{co-evolves} with the student LLM (i.e., policy model). It is continually updated in a \textbf{minimax} game. \\
\bottomrule
\end{tabular}
\vspace{0.5em}
\caption{How to implement GAD within reinforcement learning frameworks.}
\label{tbl:rl_vs_gad}
\end{table}

\paragraph{Pseudocode of Training Algorithm}
Algorithm~\ref{alg:gad} presents the pseudocode for \ours{} training.

\begin{algorithm}
\small
\caption{\ours{}: Generative Adversarial Distillation}
\label{alg:gad}
\begin{algorithmic}
\Require Distillation data $\mathcal{T} = \{(x, y_t)\}$; Student LLM (generator) $G$; Discriminator $D$
\Ensure Trained student model $G$
\Statex
\State \textbf{\textcolor{DarkBlue}{\textit{Warmup Stage}}}
\For{each batch $(x, y_t) \sim \mathcal{T}$}
    \State Update generator $G$ with cross-entropy loss on $y_t$
    \State Update discriminator $D$ with Bradley-Terry loss \Comment{~\Cref{eq:discriminator}}
\EndFor
\Statex
\State \textbf{\textcolor{DarkBlue}{\textit{\ours{} Training Stage}}}
\Repeat
    \For{each batch $(x, y_t) \sim \mathcal{T}$}
        \State Sample student responses $G(x)$
        \State Update generator $G$ using $D(G(x))$ as reward for reinforcement learning
        \State Update discriminator $D$ with Bradley-Terry loss\Comment{~\Cref{eq:discriminator}}
    \EndFor
\Until{convergence} \Return $G$
\end{algorithmic}
\end{algorithm}

\newpage
\section{Experiments}

\subsection{Setup}

\paragraph{Dataset}
Given a dataset of instruction prompts, we collect corresponding responses from a teacher model and use them to distill student models.
For the following experiments, we use \texttt{LMSYS-Chat-1M-Clean}\footnote{~\url{https://huggingface.co/datasets/OpenLeecher/lmsys_chat_1m_clean}}, a clean version of the LMSYS-Chat-1M dataset~\citep{lmsys}. The dataset is derived from high-quality conversational data collected via the Chatbot Arena\footnote{~\url{https://lmarena.ai}} platform.

\paragraph{Teacher and Student Models}
We adopt GPT-5-Chat~\citep{gpt5} as the teacher model. It is a closed-source chat model ranked ninth on the Chatbot Text Arena leaderboard at the time of writing. For student models, we use the instruction-tuned variants of open-source models from the Qwen2.5~\citep{qwen2.5} family (Qwen2.5-3B-Instruct, Qwen2.5-7B-Instruct, Qwen2.5-14B-Instruct) and the Llama3~\citep{llama3} family (Llama-3.2-3B-Instruct, Llama-3.1-8B-Instruct).

\paragraph{Training}
For training data, we sample 200K samples from \texttt{LMSYS-Chat-1M-Clean} and collect the corresponding GPT-5-Chat responses to the instructions as teacher responses. All models are trained for 3 epochs with a batch size of 256, totaling approximately 2400 optimization steps. The PPO mini-batch size for each policy update is also 256. The maximum context length is set to 2048 tokens for instruction prompts and 1536 for model responses. The training temperature is set to $0.8$. We save checkpoints every 50 steps. More training details can be found in Appendix~\ref{app:training_detail}.

\paragraph{Evaluation}
We reserve 500 samples of \texttt{LMSYS-Chat-1M-Clean} as the primary test set. We also include test datasets consisting of a 500-sample subset split from Dolly~\citep{dolly}, the 252-sample SelfInst dataset~\citep{self_inst}, and the 80-question Vicuna benchmark~\citep{vicuna} to evaluate out-of-distribution generalization.
We report the GPT-4o evaluation scores~\citep{mtbench,minillm}, where GPT-4o first generates reference answers and then scores the output of the student model against them. We also conduct human evaluations on the \texttt{LMSYS-Chat-1M-Clean} test set for qualitative assessment. 
We select the checkpoint that achieved the highest GPT-4o score and whose response length is within an acceptable range for each experiment.
Detailed evaluation protocols are described in Appendix~\ref{app:eval_detail}.

\subsection{Main Results}

\begin{table}[t]
\centering
\small
\begin{tabular}{cc|c|c|c|c}
\toprule
Model & Method
& LMSYS & Dolly & SelfInst & Vicuna \\ \midrule
 GPT-5-Chat & Teacher & 51.7 & 49.8 & 49.7 & 49.9 \\ \midrule    
 \multirow{3}{*}{Qwen2.5-3B-Instruct} 
    & Before Distill. & 45.8 & 45.1 & 45.6 & 47.3 \\
 & SeqKD  &  47.5 & 44.8 & 45.7 & 48.0 \\
 & \textbf{\ours}  &  \textbf{48.9}  &  \textbf{46.7} & \textbf{47.7} & \textbf{49.4} \\ \midrule
 \multirow{3}{*}{Qwen2.5-7B-Instruct} 
    & Before Distill. & 48.7 & 47.6 & 48.3 & 49.1 \\
 & SeqKD & 49.2 & 47.2 & 48.3 & 49.5 \\
 & \textbf{\ours} & \textbf{50.8} &  \textbf{48.5} & \textbf{50.1} &  \textbf{51.4} \\ \midrule
 \multirow{3}{*}{Qwen2.5-14B-Instruct} 
    & Before Distill. & 50.0 & 49.1 & 49.4 & 50.0 \\ 
 & SeqKD & 50.6 & 48.2 & 49.4 & 49.7 \\
 & \textbf{\ours} & \textbf{52.1} & \textbf{50.4} &  \textbf{51.1} & \textbf{51.6} \\ \midrule
 \multirow{3}{*}{Llama-3.2-3B-Instruct}                                   & Before Distill. & 44.0 & 45.8 & 47.0 & 46.9 \\
 & SeqKD & 47.6 & 47.0 & 47.1 & 48.1 \\
 & \textbf{\ours} & \textbf{48.1}  & \textbf{48.5} &  \textbf{49.1} & \textbf{48.9} \\ \midrule
 \multirow{3}{*}{Llama-3.1-8B-Instruct}                                   & Before Distill. & 46.9 & 46.6 & 48.4 & 47.9 \\
 & SeqKD & 49.7 & 47.7 & 48.7 & 48.7 \\
 & \textbf{\ours} & \textbf{50.3} & \textbf{48.8} & \textbf{49.5} & \textbf{50.2} \\
 \bottomrule
\end{tabular}
\vspace{0.4cm}
\caption{Automatic evaluation results. We report averaged GPT-4o score on the test datasets. The best results are highlighted in bold. \ours{} consistently outperforms both the instruct model before distillation and SeqKD across all datasets and model variants, with particularly strong gains in out-of-distribution generalization evaluations.}
\vspace{-0.5cm}
\label{tab:auto}
\end{table}

\label{sec:auto}

\paragraph{Automatic Evaluation}
We report the results of automatic evaluation using GPT-4o scores in \Cref{fig:scaling_student} and \Cref{tab:auto}. We compare \ours{} with the instruct model before distillation and the SeqKD baseline.
Across all datasets, \ours{} consistently outperforms the baselines. As shown in \Cref{fig:scaling_student}, on the LMSYS-Chat test set, Qwen2.5-3B-Instruct trained with \ours{} matches the performance of Qwen2.5-7B-Instruct trained with SeqKD; similarly, Qwen2.5-7B-Instruct with \ours{} rivals Qwen2.5-14B-Instruct with SeqKD, and Qwen2.5-14B-Instruct with \ours{} is comparable to the GPT-5-Chat teacher.
In addition, \ours{} shows particularly strong gains on out-of-distribution generalization benchmarks. On Dolly, SelfInst, and Vicuna, SeqKD yields marginal or even negative improvements, whereas \ours{} maintains robust performance gains. We attribute this to the superior generalization ability of reinforcement learning compared to supervised fine-tuning~\citep{sftmemrlgen,dft}. We also provide additional automatic evaluation results in \Cref{app:add_results}.

\paragraph{Human Evaluation}

\begin{figure}
\centering
\includegraphics[width=\linewidth]{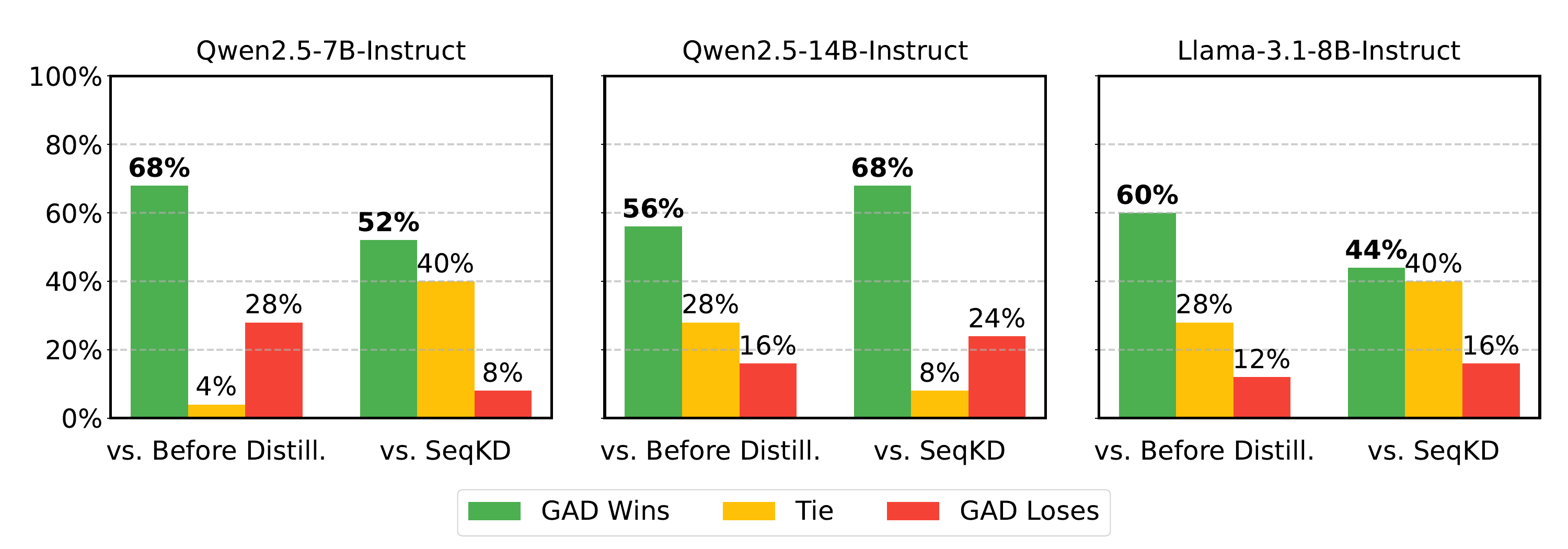}
\vspace{-0.1cm}
\caption{Human evaluation results on the \texttt{LMSYS-Chat-1M-Clean} test set. We compare \ours{} to the instruct model before distillation and the model fine-tuned with SeqKD.}
\vspace{-0.1cm}
\label{fig:user-study}
\end{figure}

We conduct human evaluations on Qwen2.5-7B-Instruct, Qwen2.5-14B-Instruct, and Llama-3.1-8B-Instruct, comparing \ours{} against both the instruct model before distillation and the model fine-tuned with SeqKD. For each prompt, the annotators assess the responses of two models and judge whether \ours{} wins, ties, or loses.
\ours{} achieves a win rate exceeding 50\% and a loss rate below 30\% in almost all comparisons. The results indicate that \ours{} can consistently outperform the baseline models on human evaluation performance.

\subsection{Analysis}
\label{sec:analysis}

\begin{figure}
    \begin{minipage}[h]{0.48\textwidth}
        \centering
        \includegraphics[width=0.9\textwidth]{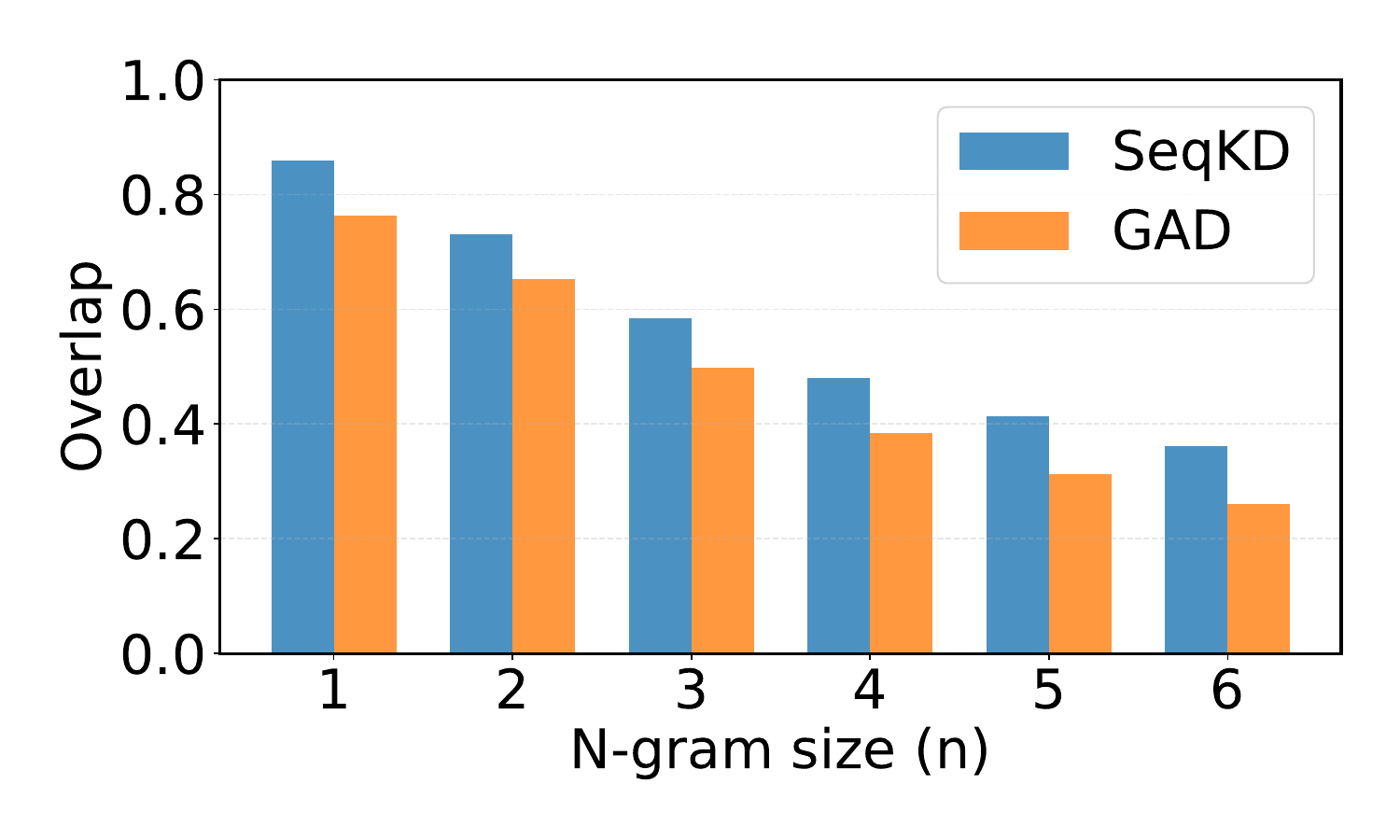}
        \caption{Overlap of local patterns between the student and the teacher. SeqKD tends to overfit to local patterns of the teacher.}
        \label{fig:ngram}
    \end{minipage}\hspace{3mm}
    \begin{minipage}[h]{0.48\textwidth}
        \centering
        \includegraphics[width=0.93\textwidth]{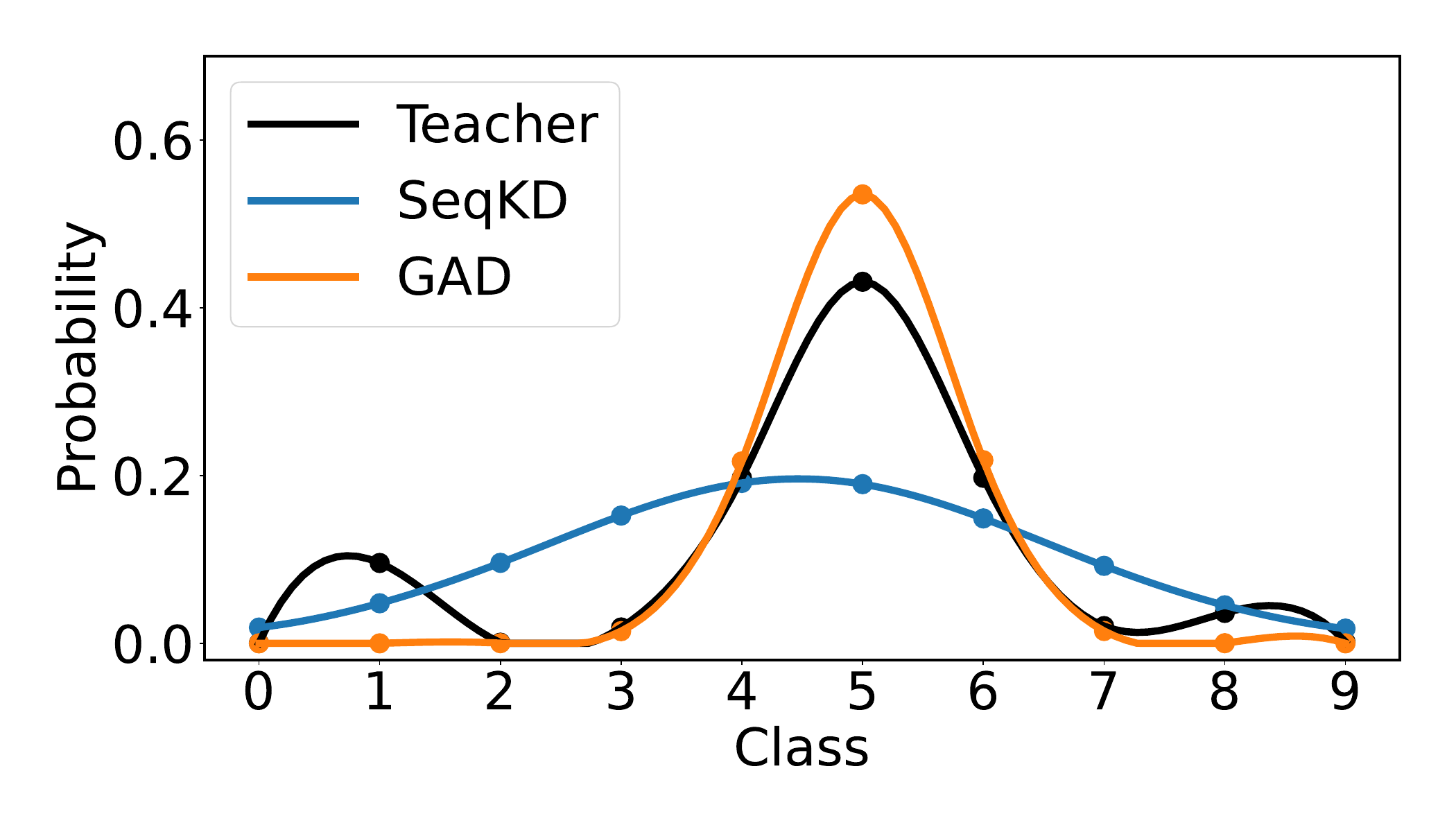}
        \caption{Black-box distillation on toy data. \ours{} learns reachable modes from the teacher while SeqKD aims to cover all the modes.}
        \label{fig:toy}
    \end{minipage}
\end{figure}

\paragraph{SeqKD Overfits to Local Patterns}

We evaluate the similarity of local patterns between the student and teacher on the LMSYS-Chat test set in \Cref{fig:ngram}, measured by the F1 score of N-gram overlap. The student is trained from Qwen2.5-14B-Instruct, and the teacher is GPT-5-Chat. The SeqKD student exhibits a higher N-gram overlap while a lower GPT-4o evaluation score compared to the \ours{} student.
This suggests that supervised fine-tuning tends to memorize local lexical patterns~\citep{sftmemrlgen,dft}, whereas our RL-based approach better captures the teacher's global stylistic characteristics.

\paragraph{Experiments on Toy Data}

We simulate the optimizing patterns of \ours{} and SeqKD in a toy experiment shown in \Cref{fig:toy}. We observe that \ours{} tends to learn reachable modes of the teacher, whereas SeqKD aims to cover all modes.
The setup simulates a black-box distillation scenario. We define a discrete Gaussian mixture distribution as a teacher distribution $p$, which has categorical outputs ${0, \dots, 9}$. A student, modeled as a single Gaussian distribution, learns to imitate the teacher using only output samples without access to $p$.
We compare two student training schemes, SeqKD and \ours{}. The \ours{} student is optimized using the REINFORCE algorithm~\citep{reinforce}.
As illustrated in \Cref{fig:toy}, the SeqKD student exhibits a mode-covering behavior, spreading probability mass across all possible outputs~\citep{minillm}. In contrast, the \ours{} student focuses on mode-seeking, concentrating probability optimization on reachable regions. We find that such mode-seeking behavior leads to more effective knowledge distillation in LLMs.

\paragraph{Comparison to Off-Policy Discriminator}

\begin{wrapfigure}{r}{7.0cm}
\centering
\vspace{-0.45cm}
\includegraphics[width=0.50\textwidth]{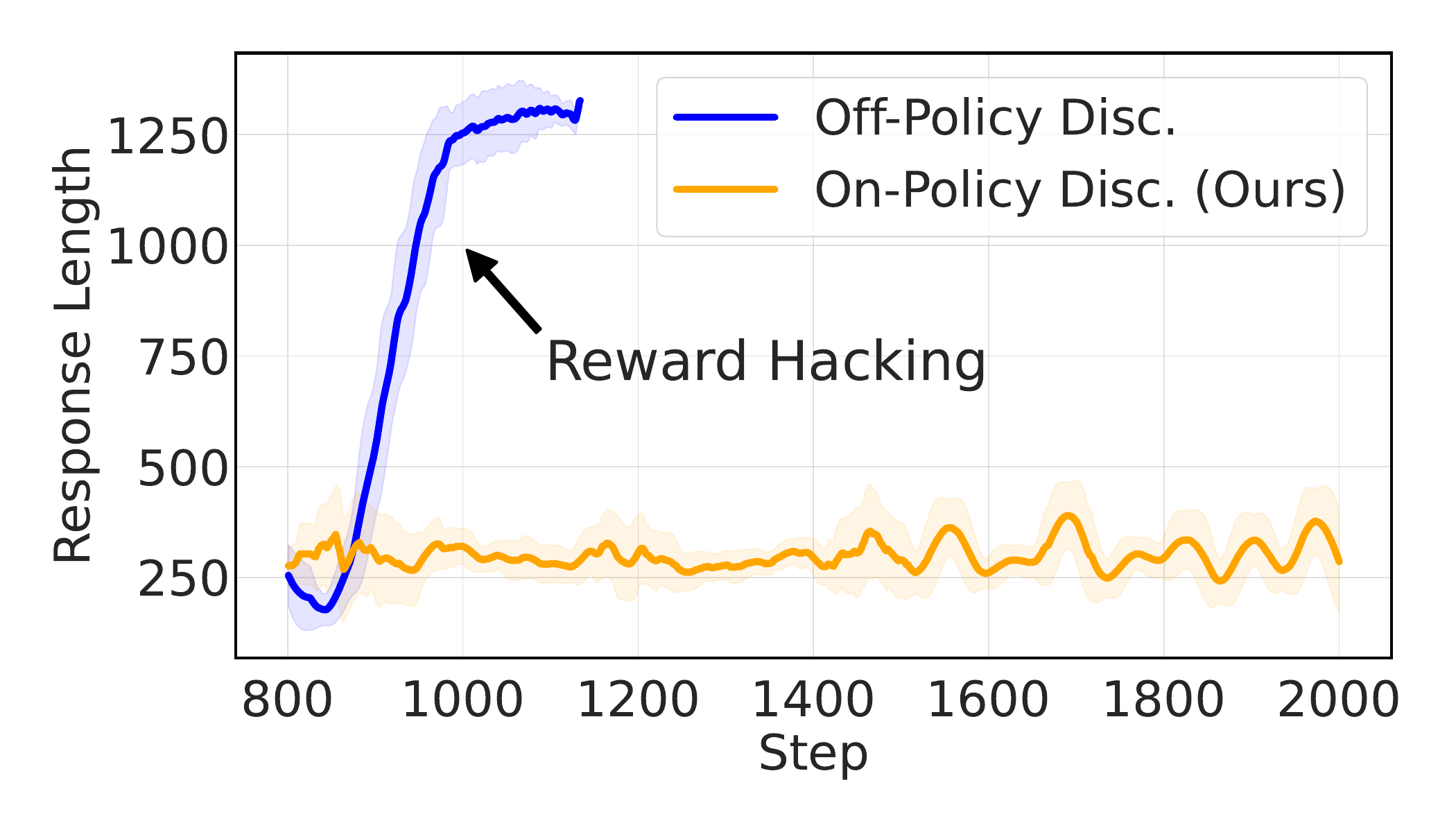}
\vspace{-0.45cm}
\caption{Off-policy discriminator suffers from reward hacking, whereas on-policy discriminator remains stable over thousands of training steps.}
\vspace{-0.1cm}
\label{fig:reward}
\end{wrapfigure}

As discussed in \Cref{sec:gad}, from the view of reinforcement learning, our generator (student) acts as the policy model, while the discriminator acts as the on-policy reward model. 
\Cref{fig:reward} compares \ours{} with the off-policy discriminator approach. In the off-policy setting, the student is first trained for one warmup epoch using SeqKD. The student is then frozen, and the discriminator is trained for two epochs based on the student's output. Then the resulting discriminator serves as a frozen reward model to train the student using \Eqref{eq:grpo-generator}.
In contrast, \ours{} jointly trains the student and discriminator for one warmup epoch followed by two \ours{} training epochs, positioning the discriminator as an on-policy reward model.
We observe that the student trained with an off-policy discriminator quickly exhibits reward hacking after around 300 training steps, producing excessively long responses (up to 1300 tokens) that deviate significantly from the teacher's patterns. In comparison, \ours{} remains stable through thousands of training steps with no sign of reward hacking. The results establish \ours{} as a highly reliable and robust on-policy distillation method.




\subsection{Ablations}
\label{sec:ablation}

\begin{figure}[t]
\begin{minipage}[h]{0.48\textwidth}
    \centering
    \small
    \begin{tabular}{l|cc}
    \toprule
      &  LMSYS & Others \\ \midrule
    SeqKD  &  49.2  &  48.3  \\  \midrule
    \ours{}  &  \textbf{50.8}  &  \textbf{50.0}  \\
    \ \ w/o Gen. Warmup  &  49.7  &  49.7  \\
    \ \ w/o Disc. Warmup  &  49.0  &  47.7  \\
    \bottomrule
    \end{tabular}
    \makeatletter\def\@captype{table}\makeatother\caption{Ablation of warmup strategy on Qwen2.5-7B-Instruct. Warmup of the generator and discriminator are removed separately.}
    \label{tab:warmup}
\end{minipage}\hspace{3mm}
\begin{minipage}[h]{0.48\textwidth}
    \centering
    \small
    \begin{tabular}{l|cc}
    \toprule
      &  LMSYS & Others \\ \midrule
    SeqKD  &  47.5   &  46.2  \\  \midrule
    \ours{}  &   &   \\
    \ \ Disc. BT Loss (Default)  &  \textbf{48.9}  &  \textbf{47.9}  \\
    \ \ Disc. CE Loss  &  47.9  &  46.4   \\
    \bottomrule
    \end{tabular}
    \makeatletter\def\@captype{table}\makeatother\caption{Ablation of discriminator loss choice on Qwen2.5-3B-Instruct. Default Bradley-Terry loss outperforms cross-entropy loss.}
    \label{tab:discloss}
\end{minipage}
\end{figure}

\paragraph{Warmup Strategy}
We perform an ablation study of the warmup strategy introduced in \Cref{sec:rl_optim}. As shown in \Cref{tab:warmup}, we separately remove the warmup stage for the generator and the discriminator on Qwen2.5-7B-Instruct.
When removing the generator warmup, we directly use Qwen2.5-7B-Instruct without SeqKD as initialization for both the generator and discriminator for \ours{} training. This leads to a performance drop. We attribute this to the discriminator easily distinguishing between the student and teacher outputs in the early training stage. The large distributional gap between the teacher and the student weakens the effectiveness of \ours{} training.
When removing the discriminator warmup, we use the generator obtained after one epoch of SeqKD and initialize the discriminator with the original Qwen2.5-7B-Instruct. In this setting, the imbalance between the generator and the discriminator prevents the discriminator from providing sufficiently informative feedback. Consequently, the adversarial interaction becomes ineffective, and the generator exhibits little improvement beyond its warmup performance.

\paragraph{Discriminator Loss Choice} 
We ablate the choice of discriminator loss in \Cref{tab:discloss}, and observe that our default Bradley-Terry loss outperforms cross-entropy loss in overall GPT-4o evaluation score. The Bradley-Terry loss is defined in \Eqref{eq:discriminator}, while the cross-entropy loss is a binary classification loss commonly adopted for discriminator in prior works~\citep{gan,seqgan,GAIL,learningdense}. The cross-entropy discriminator loss can be written as:

\begin{equation}
\label{eq:ce-discriminator}
\min_D \ \mathbb{E}_{(x,y_t) \sim \mathcal{T}} \left[-\log \sigma \left( D(y_t) \right) - \log \left( 1 - \sigma \left( D(G(x)) \right) \right) \right].
\end{equation}

Experiments on Qwen2.5-3B-Instruct shows that Bradley-Terry loss can enhance discriminator training stability and improve automatic evaluation scores over cross-entropy loss. The result highlights the effectiveness of Bradley-Terry loss for discriminator training in LLMs.

\begin{wrapfigure}{r}{6.7cm}
\centering
\begin{tabular}{@{}cc|cc@{}}
\toprule
Gen. Size &  Disc. Size & LMSYS & Others \\ \midrule
3B & 3B (Default)  &  \textbf{48.9}  &  \textbf{47.9} \\
3B & 7B  &  47.8 &  46.9  \\   \midrule
7B & 7B (Default) &  \textbf{50.8}  &  \textbf{50.0}   \\
7B & 14B  &  50.5  &  49.9   \\
\bottomrule
\end{tabular}
\makeatletter\def\@captype{table}\makeatother\caption{Ablation of discriminator model size. Experiments are performed on Qwen2.5 Instruct models, evaluated with GPT-4o score.}
\label{tab:discsize}
\vspace{-1em}
\end{wrapfigure}

\paragraph{Discriminator Model Size}

We ablate the relative model size of the generator and discriminator in \Cref{tab:discsize}. Using equal model sizes for the two models, which is our default setting, yields the best performance. The experiments are conducted on Qwen2.5 Instruct models and evaluated by GPT-4o scores. We increase the discriminator model size from 3B to 7B for the 3B student, and from 7B to 14B for the 7B student in \ours{}. Increasing the discriminator size does not improve performance. The experiment shows maintaining a balanced generator-discriminator pair is crucial for achieving strong performance.

\section{Related Work}

\paragraph{White-box Distillation of LLM} 
White-box knowledge distillation of LLM assumes full access to the internal representations or token-level probabilities of a teacher model.
Standard white-box approaches align the forward KLD of distribution~\citep{mixkd,lightpaff}, reverse KLD of distribution~\citep{minillm}, hidden states~\citep{tinybert,bert-pkd} or attention scores~\citep{minilm,minilmv2} between the teacher and the student. Recent work~\citep{minillm,thinkingmachine-onpolicy,googlepolicy} also proves the importance of on-policy distillation where the student learns from its own responses.
Such approaches effectively compress large models while preserving semantic similarity. Despite their effectiveness, these methods rely on full teacher access, which is impractical for proprietary LLMs and limits their applicability to closed-source or API-only teachers.

\paragraph{Black-box Distillation of LLM}
Black-box distillation trains a student model using only the textual outputs of a teacher, typically obtained by API queries to closed-source models such as GPT-5 and Gemini 2.5~\citep{gpt5,gemini2d5}.
In this setting, conventional white-box distillation methods become infeasible because of the lack of access to the teacher's logits or hidden representations.
The standard approach for this scenario, SeqKD, performs supervised fine-tuning (SFT) on the teacher's responses~\citep{skd,ITGPT4,lima,alpaca,vicuna} to imitate the teacher's behaviors.
Recent work~\citep{s1,openthoughts,limo,deepseekr1} extends this paradigm by performing SFT on the teacher's reasoning traces to improve the student's reasoning ability.

\section{Conclusion}

We introduce \ours{}, a generative adversarial framework that effectively addresses key challenges of black-box LLM distillation. \ours{} enables on-policy learning by training a student model and an adaptive discriminator in a minimax game, eliminating the need for any logit-level supervision. This discriminator provides an implicit, on-policy reward signal that guides the student's optimization. Experiments across multiple model families and datasets confirm our approach. \ours{} consistently surpasses standard sequence-level distillation, delivering superior generalization and achieving performance that rivals the proprietary teacher. These results validate \ours{} as an effective and robust solution for black-box LLM distillation.

\section*{Acknowledgements}

We are grateful to Yi Zhu for technical support during the development of the RL infrastructure and to Yuxian Gu for insightful discussions.

\bibliographystyle{alpha}
\bibliography{neurips_2023}

\newpage
\appendix

\section{Experimental Details}

\subsection{Implement \ours{} with GRPO}
\label{app:grpo}

We implement policy optimization of the student with GRPO~\citep{grpo}. We use $q_{G}$ to denote the output distribution of student $G$. For each input prompt $x$, we sample a group of $N$ student responses $\{y_s^i\}_{i=1}^N$, and obtain their corresponding rewards $\{r_s^{i}\}_{i=1}^N$, where $r_s^{i} = D(y_s^i)$.
The advantage of the $i$-th response can be calculated with:
\begin{align}
\label{eq:grpo-advantage}
r_s^{i} &= D(y_s^i) \\
A^i &= \frac{r_s^i - \text{mean}(\{r_s^j\}_{j=1}^N)}{\text{std}(\{r_s^j\}_{j=1}^N)}.
\end{align}
The student is trained with the following objective:
\begin{align}
\label{eq:grpo-generator}
\max_{G} \ \ & \mathbb{E}_{(x,y_t) \sim \mathcal{T}, \{y_s^i\}_{i=1}^N \sim q_{G}(\cdot \mid x)} \left[ \frac{1}{N} \sum_{i=1}^N A^i \right],
\end{align}
where we omit the KL regularizer and the clip operator in GRPO for brevity.

For the discriminator, we pair each student response $y_s^i$ in the group with the same teacher response $y_t$ to form $(y_t, y_s^i)$ preference pairs.
The discriminator parameters are optimized by minimizing the Bradley-Terry loss across the group:
\begin{equation}
\label{eq:grpo-discriminator}
\min_D \ \ \mathbb{E}_{(x,y_t) \sim \mathcal{T}, \{y_s^i\}_{i=1}^N \sim q_{G}(\cdot \mid x)} \left[ \frac{1}{N} \sum_{i=1}^N -\log \sigma(D(y_t) - D(y_s^i)) \right],
\end{equation}
where $D(y_t)$ is the teacher score shared within the group.

\subsection{Training Details}
\label{app:training_detail}
We train all models with 3 epochs. For \ours{}, the training consists of 1 warmup epoch followed by 2 \ours{} training epochs. 
The models are trained with a batch size of 256, totaling approximately 2400 optimization steps. The PPO mini-batch size for each policy update is also 256. In the warmup stage of \ours{}, we train the discriminator for 10 steps before jointly training the generator and discriminator.

We search learning rate in [1e-6, 5e-6] for \ours{} and SeqKD baseline. For SeqKD, we find 5e-6 leads to better results in all experiments. For \ours{} with GPT-5-Chat teacher, we use 1e-6 for both warmup and \ours{} training stage, and for \ours{} with Qwen2.5 teacher as in \Cref{tab:auto-extend-2}, we use 5e-6 for warmup stage and 1e-6 for \ours{} training stage. The maximum context length is set to 2048 tokens for instruction prompts and 1536 for model responses. The training temperature is set to $0.8$.

In the GRPO algorithm formulated as \Eqref{eq:grpo-generator}, we set group size $N=8$ and the KL weight $\beta=0.001$.

Distilling Qwen2.5-14B-Instruct from GPT-5-Chat takes about 30 hours on 16 H100 GPUs.

\newpage
\subsection{Automatic Evaluation Details}
\label{app:eval_detail}
We use greedy sampling and the model response length is set to 1536 tokens.
We use the prompt wrapper in Figure \ref{fig:prompt_wrapper} to construct prompts.
We use the prompt in Figure \ref{fig:prompt_gpt4} for GPT-4o feedback following~\citep{minillm}. The reported GPT-4o score is defined as the student's score divided by the sum of the student's score and the reference answer's score.

\begin{figure}[h]
    \begin{tcolorbox}
    Below is an instruction that describes a task. \\
    Write a response that appropriately completes the request. \\ \\
    \#\#\# Instruction: \\
    \{instruction\} \\ \\
    \#\#\# Response:
    \end{tcolorbox}
    \caption{The prompt wrapper for training and evaluation.}
    \label{fig:prompt_wrapper}
\end{figure}

\begin{figure}[h]
    \begin{tcolorbox}
        We would like to request your feedback on the performance of two AI assistants in response to the user instruction and input displayed above. \\
        Please rate the helpfulness, relevance, accuracy, and level of detail of their responses. Each assistant receives an overall score on a scale of 1 to 10, where a higher score indicates better overall performance. \\
        Please first output a single line containing only two values indicating the scores for Assistant 1 and 2, respectively. The two scores are separated by a space. \\
        In the subsequent line, please provide a comprehensive explanation of your evaluation, avoiding any potential bias and ensuring that the order in which the responses were presented does not affect your judgment.
    \end{tcolorbox}
    \caption{GPT-4o evaluation prompt.}
    \label{fig:prompt_gpt4}
\end{figure}
















\section{Additional Results}

\subsection{Additional Automatic Evaluation Results}
\label{app:add_results}

\paragraph{GPT-5 Teacher}

We provide additional results of the automatic evaluation.
In \Cref{tab:auto-extend-1}, we report GPT-4o score and response lengths of distilled student models trained with the GPT-5-Chat teacher. Across datasets, we observe that SeqKD tends to produce shorter responses that closely follow the teacher's length distribution whereas \ours{} maintains the original model's length distribution while integrating the teacher's global stylistic characteristics. We attribute this behavior to the on-policy sampling of \ours{}, which encourages generation patterns aligned with both the student's prior and the teacher's guidance.

\begin{table}[t]
    \centering
    \small
    \begin{tabular}{cc|cc|cc|cc|cc}
    \toprule
    \multirow{2}{*}{Model} & \multirow{2}{*}{Method}
    & \multicolumn{2}{c|}{LMSYS} & \multicolumn{2}{c|}{Dolly} & \multicolumn{2}{c|}{SelfInst} & \multicolumn{2}{c}{Vicuna} \\
     & & Score & Len. & Score & Len. & Score & Len. & Score & Len. \\ \midrule
     GPT-5-Chat & Teacher & 51.7 & 329.1 & 49.8 & 148.5 & 49.7 & 188.5 & 49.9 & 378.6 \\ \midrule    
     \multirow{3}{*}{Qwen2.5-3B-I} 
        & Before Distill. & 45.8 & 338.9 & 45.1 & 219.2 & 45.6 & 279.3 & 47.3 & 520.9 \\
     & SeqKD  &  47.5 & 318.2 & 44.8 & 160.6 & 45.7 & 207.1 & 48.0 & 370.4 \\
     & \textbf{\ours}  &  \textbf{48.9}  &  438.0 &  \textbf{46.7} & 239.5 & \textbf{47.7} & 281.8 & \textbf{49.4} & 517.9 \\ \midrule
     \multirow{3}{*}{Qwen2.5-7B-I} 
        & Before Distill. & 48.7 & 345.2 & 47.6 & 220.0 & 48.3 & 259.1 & 49.1 & 501.7 \\
     & SeqKD & 49.2 & 320.2 & 47.2 & 152.3 & 48.3 & 182.3 & 49.5 & 398.1 \\
     & \textbf{\ours} & \textbf{50.8} & 414.0 &  \textbf{48.5} & 225.1 & \textbf{50.1} & 288.5 &  \textbf{51.4} & 511.9 \\ \midrule
     \multirow{3}{*}{Qwen2.5-14B-I} 
        & Before Distill. & 50.0 & 322.1 & 49.1 & 201.6 & 49.4 & 252.0 & 50.0 & 475.4 \\ 
     & SeqKD & 50.6 & 319.3 & 48.2 & 151.2 & 49.4 & 199.8 & 49.7 & 402.5 \\
     & \textbf{\ours} & \textbf{52.1} & 438.9 & \textbf{50.4} & 262.6 &  \textbf{51.1} & 284.1 & \textbf{51.6} & 499.6 \\ \midrule
     \multirow{3}{*}{Llama-3.2-3B-I}                                                 & Before Distill. & 44.0 & 334.4 & 45.8 & 174.5 & 47.0 & 265.6 & 46.9 & 437.6 \\
     & SeqKD & 47.6 & 328.6 & 47.0 & 147.4 & 47.1 & 214.5 & 48.1 & 389.3 \\
     & \textbf{\ours} & \textbf{48.1}  & 371.5 & \textbf{48.5} & 232.3 &  \textbf{49.1} & 275.7 & \textbf{48.9} & 461.8 \\ \midrule
     \multirow{3}{*}{Llama-3.1-8B-I}                                                 & Before Distill. & 46.9 & 329.2 & 46.6 & 184.7 & 48.4 & 276.2 & 47.9 & 487.8 \\
     & SeqKD & 49.7 & 319.6 & 47.7 & 148.4 & 48.7 & 199.7 & 48.7 & 400.3 \\
     & \textbf{\ours} & \textbf{50.3} & 394.6 & \textbf{48.8} & 200.6 &  \textbf{49.5} & 263.8 & \textbf{50.2} & 504.2 \\
     \bottomrule
    \end{tabular}
    \vspace{0.4cm}
    \caption{Extended automatic evaluation results with GPT-5-Chat teacher. We report averaged GPT-4o score and token length of response.}
    \vspace{-0.5cm}
    \label{tab:auto-extend-1}
\end{table}

\begin{table}[h]
    \centering
    \small
    \begin{tabular}{cc|c|c|c|c}
    \toprule
    Model & Method
    & LMSYS & Dolly & SelfInst & Vicuna \\ \midrule
     Qwen2.5-14B-I & Teacher & 50.0 & 49.1 & 49.4 & 50.0 \\ \midrule
     \multirow{3}{*}{Llama-3.2-3B-I}                                                 & Before Distill. & 44.0 & 45.8 & 47.0 & 46.9 \\
     & SeqKD & 46.9 & 47.6 & \textbf{47.6} & 48.5 \\
     & \textbf{\ours} & \textbf{47.5} & \textbf{47.7} & 47.3 & \textbf{49.0} \\ \midrule
     \multirow{3}{*}{Llama-3.1-8B-I}                                                 & Before Distill. & 46.9 & 46.6 & 48.4 & 47.9 \\
     & SeqKD & 49.0 & 48.4 & 48.6 & 49.4 \\
     & \textbf{\ours} & \textbf{49.6} & \textbf{49.9} & \textbf{50.5} & \textbf{49.7} \\
     \bottomrule
    \end{tabular}
    \vspace{0.4cm}
    \caption{Automatic evaluation results with Qwen2.5-14B-Instruct teacher. We report averaged GPT-4o score.}
    \vspace{-0.5cm}
    \label{tab:auto-extend-2}
\end{table}

\paragraph{Qwen2.5 Teacher}
In \Cref{tab:auto-extend-2}, we distill from Qwen2.5-14B-Instruct teacher to student models from the Llama family. Although the teacher is open-source, its tokenizer is incompatible with the students, preventing direct application of white-box distillation methods that align KL divergence between teacher and student logits. In this setting, \ours{} remains effective, outperforming both the pre-distillation models and the SeqKD baseline in most settings on GPT-4o evaluation score.

\end{document}

%% file: math_commands.tex
\usepackage{amsmath,amsfonts,bm}

\def\eqref#1{equation~(\ref{#1})}
\def\Eqref#1{Equation~(\ref{#1})}

\def\1{\bm{1}}

\DeclareMathAlphabet{\mathsfit}{\encodingdefault}{\sfdefault}{m}{sl}
\SetMathAlphabet{\mathsfit}{bold}{\encodingdefault}{\sfdefault}{bx}{n}



%% file: settings.tex
\usepackage{multirow}
\usepackage{amsmath}
\usepackage{capt-of}
\usepackage{tabularx}
\usepackage{epsfig}
\usepackage{amssymb}
\usepackage{amsfonts}
\usepackage{booktabs}
\usepackage{scalerel}
\usepackage{listings}
\usepackage{varwidth}
\usepackage{stmaryrd}
\usepackage{bbm}
\usepackage{wrapfig}
\usepackage{pifont}

\newcommand{\tabincell}[2]{\begin{tabular}{@{}#1@{}}#2\end{tabular}}

\definecolor{deepblue}{rgb}{0,0,0.5}
\definecolor{officeblue}{RGB}{0,102,204}
\definecolor{deepred}{rgb}{0.6,0,0}
\definecolor{deepgreen}{rgb}{0,0.5,0}
\definecolor{mybrickred}{RGB}{182,50,28}

\definecolor{fillcolor}{RGB}{216,217,252}

\usepackage{etoolbox}
\usepackage{framed}

\newif\ifxetexorluatex
\ifxetex
  \xetexorluatextrue
\else
  \ifluatex
    \xetexorluatextrue
  \else
    \xetexorluatexfalse
  \fi
\fi
\newcommand*\quotesize{60} 
\newcommand*{\openquote}
   {\tikz[remember picture,overlay,xshift=-4ex,yshift=-2.5ex]
   \node (OQ) {\fontsize{\quotesize}{\quotesize}\selectfont``};\kern0pt}

\newcommand*{\closequote}[1]
  {\tikz[remember picture,overlay,xshift=4ex,yshift={#1}]
   \node (CQ) {\fontsize{\quotesize}{\quotesize}\selectfont''};}

\colorlet{shadecolor}{white}

\newcommand*\shadedauthorformat{\emph} 

\newcommand*\authoralign[1]{%
  \if#1l
    \def\authorfill{}\def\quotefill{\hfill}
  \else
    \if#1r
      \def\authorfill{\hfill}\def\quotefill{}
    \else
      \if#1c
        \gdef\authorfill{\hfill}\def\quotefill{\hfill}
      \else\typeout{Invalid option}
      \fi
    \fi
  \fi}
{\authoralign{#1}
\ifblank{#2}
   {\def\shadequoteauthor{}\def\yshift{-2ex}\def\quotefill{\hfill}}
   {\def\shadequoteauthor{\par\authorfill\shadedauthorformat{#2}}\def\yshift{2ex}}
\begin{snugshade}\begin{quote}\openquote}
{\shadequoteauthor\quotefill\closequote{\yshift}\end{quote}\end{snugshade}}
